\documentclass[letterpaper, 10 pt, conference]{ieeeconf}  

\IEEEoverridecommandlockouts                              

\usepackage{graphics} 
\usepackage{epsfig} 
\usepackage{mathptmx} 
\usepackage{times} 
\usepackage{amsmath} 
\usepackage{amssymb}  

\usepackage{times}
\usepackage[font=footnotesize, skip=10pt]{caption}
\usepackage{multicol}
\usepackage{multirow}
\usepackage[bookmarks=true]{hyperref}
\usepackage[utf8]{inputenc}
\usepackage{amsmath,amssymb}
\usepackage{algorithm}
\usepackage{algorithmic}
\usepackage{graphicx}
\usepackage{booktabs}
\usepackage{xcolor}
\usepackage{xspace}
\usepackage{verbatim}
\usepackage{siunitx} 
\usepackage{algorithm}
\usepackage{algorithmic}
\usepackage{cuted}   

\newcommand{\method}{PREFAIL\xspace}

\makeatletter
\let\origthebibliography\thebibliography
\def\thebibliography#1{%
  \origthebibliography{#1}%
  \scriptsize  
}
\makeatother

\title{\LARGE \bf \method: Identifying Precursors to Failures in Robotic \\ Lift-and-Place Tasks to Improve Task Execution Performance
}

\author{
Zeyu Shangguan, Rajas Chitale, Rutvik Patel, Satyandra K. Gupta and Daniel Seita
\thanks{Thomas Lord Department of Computer Science, University of Southern
California, USA. Correspondence to: {\tt\small zshanggu@usc.edu} and {\tt\small seita@usc.edu.}}%
}

\begin{document}

\maketitle
\thispagestyle{empty}
\pagestyle{empty}

\begin{abstract}

Non-prehensile manipulation enables flexible material handling with part  carriers, but friction-based support makes high-speed motions failure-prone, while slower operation increases cycle time. Proactive failure prediction is therefore essential for efficient and reliable performance, yet existing approaches remain limited by key constraints, including sensitivity to dynamic actions and high dependence on known policy structures. Furthermore, existing methods and datasets lack a precise characterization of the latest intervention time, leaving it unclear whether a detected failure can still be prevented through timely intervention. 
In this paper, we investigate lift-and-place tasks for non-prehensile material handling manipulation and propose a more effective approach to predicting precursors to failures (\method) by analyzing the relative motion of target objects with respect to the carrier. We further introduce a dataset that precisely identifies the latest intervention time for risky manipulations, enabling rigorous evaluation of whether a failure prediction is actionable. We validate our approach on both simulation and real-world datasets. Our experimental results demonstrate that \method substantially improves both the accuracy and timeliness of responses to failure precursors.
Our website with supplementary material is available at: https://zshanggu.github.io/zeyu-prefail/
\end{abstract}

\section{INTRODUCTION}


Non-prehensile manipulation in industrial material handling application~\cite{mason1986mechanics, lynch1996stable, 8280543}, such as pushing, sliding, tilting, or sweeping, unlike grasping, relies solely on contact forces to support objects, where rapid direction changes introduce large inertial forces that can overcome friction, leading to failure, object damage, or costly human intervention. In contrast, operating at uniformly low speeds is impractical due to high cycle times. Therefore, it is highly desirable to have a failure prediction system that enables high-speed operation while continuously monitoring risk to achieve fast cycle times with reduced failure risk.

Current failure prediction methods demonstrate success in controlled laboratory settings~\cite{ICLR2025_70a06501, thoduka2024multimodal, lemasurier2024reactive}, however, we observe that predicting failures during high-speed non-prehensile robot manipulation remains challenging. First, training data for failure cases during dynamic manipulation are rare. Second, we observe that current out-of-distribution (OOD) detection methods~\cite{XuC2-RSS-25} struggle to distinguish between failure and non-failure trajectories in rapid manipulation scenarios. Moreover, many existing approaches hinge on predicting future actions from a learned policy. This design tightly couples failure detection to a specific control strategy, limiting cross-scenario generalization. 
These limitations motivate us to explore a novel approach that is (i) policy-agnostic and (ii) adapt to fast-moving non-prehensile manipulation scenario. In this paper, we focus on rapid lift-and-place as a representative task, which is widely used in non-prehensile material handling applications.


\begin{figure}[t]
    \centering
    \includegraphics[width=\linewidth]{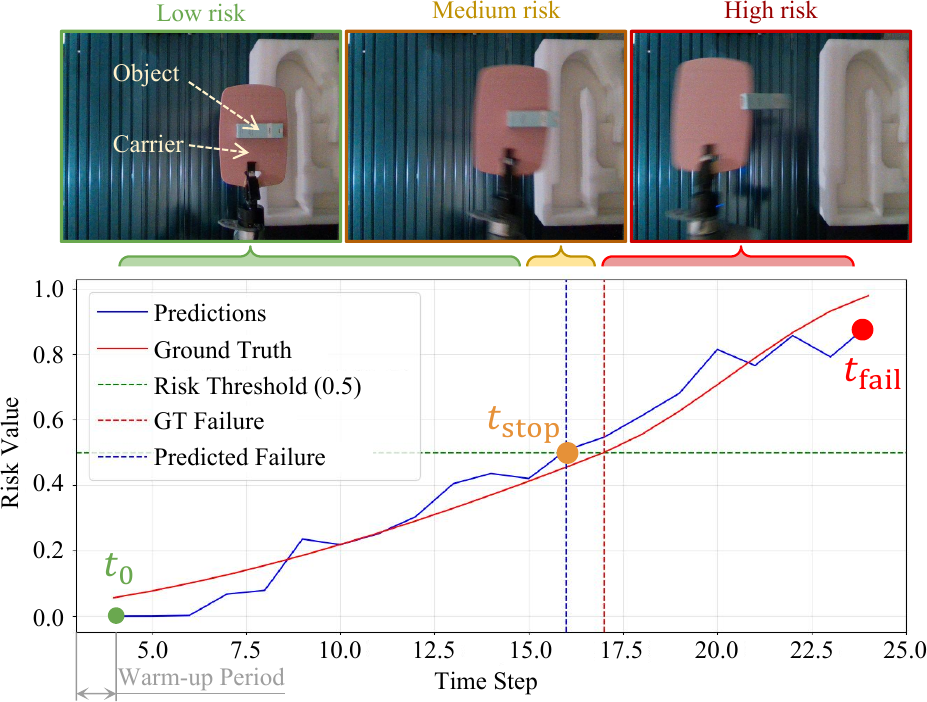}
    \caption{Our lift-and-place task involves a carrier (the pink plate in the image) mounted on a robot arm that handles an object (the cyan box) at the maximum allowable speed. Our method utilizes the visual observation and robot state values during a real-time robotic lift-and-place task, and predicts a continuous risk value (blue solid line) that represents the likelihood of failure. Once the risk is approaching the threshold (medium risk zone), the system must adjust and slow its speed, otherwise, the failure is inevitable after $t_\text{stop}$ (high risk). Since we will use sliding window during inference, there will be a warm-up period for collecting sufficient history information.}
    \label{fig:intro}
    \vspace{-20pt}
\end{figure}

We hypothesize that, under fast manipulation settings, failure prediction systems must react rapidly to scene changes. Drawing an analogy from human driving: at highway speeds, drivers assess risk from surrounding vehicles’ motion and adjust accordingly, but braking too late makes a collision unavoidable. This intuition suggests two key insights for robotic manipulation. First, for non-prehensile lift-and-place, unusual relative motion between objects can serve as an early indicator of impending failure. Second, there exists a critical time threshold beyond which even standard emergency intervention on industrial robots, such as emergency stop, cannot prevent failure due to inertia and contact dynamics.
Existing benchmarks for failure prediction~\cite{mandlekar2022matters, jia2024towards, chi2025diffusion, agia2025unpacking} do not capture this critical temporal boundary, making it difficult to evaluate whether a system predicts failures early enough for effective intervention. To address this gap, we propose a novel data collection pipeline that precisely labels the latest moment at which an emergency stop can still prevent failure (denoted as $t_\text{stop}$, see Sec.~\ref{sec:method}). This labeled threshold enables us to evaluate whether our method predicts failures with sufficient notice for real-time corrective action. 

Motivated by these insights, we propose \method, a policy-agnostic failure prediction framework for fast lift-and-place manipulation. \method monitors geometric changes in the scene and predicts whether failures will occur before the critical deadline $t_\text{stop}$, enabling timely intervention.
\method utilizes a two-branch architecture. 
The observation branch extracts multi-view visual features and relative object-carrier motion, and a state branch encodes proprioceptive signals and past risk estimates. These features are fused to predict a continuous risk value in $[0,1]$. 
Our goal is to correctly predict the boundary risk value before $t_\text{stop}$ to enable timely interventions to avoid an irrecoverable failure.

We evaluate \method using both simulation and real-world data, comparing against a competitive OOD baseline~\cite{XuC2-RSS-25}. \method achieves superior performance across multiple metrics: precision, accuracy and recall for correct failure identification, and intervention timeliness ratio, a metric we introduce that measures the percentage of manipulation completed before triggering an intervention (see Sec.~\ref{sec:assessment}). A high intervention timeliness ratio indicates the system avoids overly conservative predictions that unnecessarily halts manipulation early, while still preventing failures.

Our contributions include:
\begin{itemize}
    \item \method, a novel framework for proactive failure prediction for rapid non-prehensile lift-and-place manipulation that analyzes the relative motion between the object and the carrier.
    \item A dataset with precise temporal labeling that identifies the critical moment beyond which even emergency interventions cannot prevent failure, enabling evaluation of prediction timeliness in addition to accuracy.
    \item Experiments in simulation and the real world demonstrating that our motion-based approach achieves higher performance compared to baselines.
\end{itemize}

\section{RELATED WORKS}




\begin{figure*}[h]
    \centering
    \includegraphics[width=\linewidth]{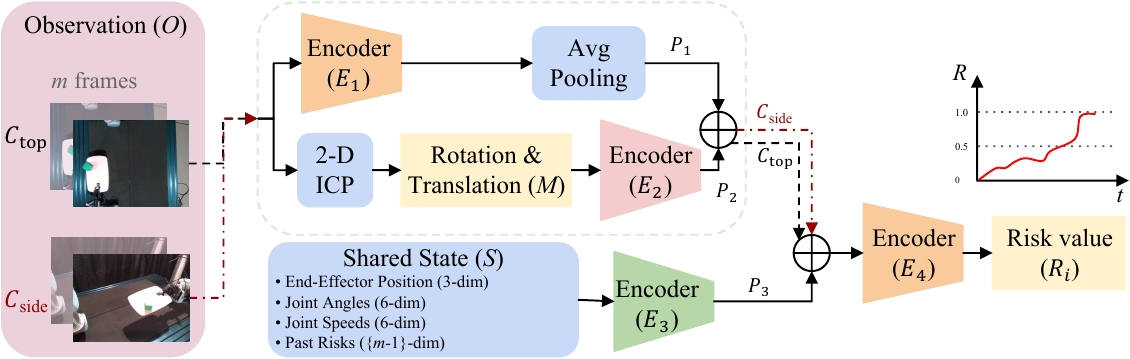}
    \caption{\textbf{Overview of \method} (see Sec.~\ref{sec:method} for details). For each sample with $m$ time steps, the observation branch processes multi-view images to extract visual features $P_1$, then uses the segmented contours of the target object and the carrier to calculate their relative motion to get the motion features $P_2$. The state branch uses $E_3$ to encode robot states and past risk values to produce features $P_3$. We fuse and process features with $E_4$ to output a continuous risk value $R_i$ indicating the likelihood of a failure at each time step $t_i$.}
    \label{fig:Pipeline}
    \vspace{-15pt}
\end{figure*}

\subsection{Failure Detection}

Current research on failure identification in robotic manipulation can be broadly categorized into two approaches: \emph{reactive analysis} and \emph{proactive prediction}.
Prior methods for reactive analysis include failure explanation using natural language~\cite{lemasurier2024reactive, agia2025unpacking, zheng2024evaluating}, closed set classification~\cite{Inceoglu2023Multimodal, ma2024multimodal}, or using Large Language Models (LLMs) to aid in failure detection~\cite{liu2023reflect,ICLR2025_70a06501,kang2024using,pramanick2024multimodal, dai2025racer, sinha2024real}. Despite progress, their post-hoc nature limits use in industrial settings, where proactive prediction and prevention are crucial. 

Compared to reactive failure detection, proactive failure prediction in robotic manipulation focuses more on real-time failure detection and prevention. Sogi~et~al.~\cite{sogi2024future} propose a proactive success-or-failure classification framework. Enshen~et~al.~\cite{zhou2025code} propose using geometric constraints for real-time failure prediction. Ping~et~al.~\cite{huang2021task} use machine learning to predict maximum object shift for wafer transferring scenarios. Parashar~et~al.~\cite{parashar2025failure} propose to combine exhaustive simulation-based falsification with Gaussian process regression to bridge the sim-to-real gap. Lee~et~al.~\cite{lee2025grasp} propose an analytical grasp failure model for multi-suction-cup grippers that solves load distribution via a minimum spring potential energy quadratic program. Sagar~et~al.~\cite{sagar2026uncovering} generates action probability distributions to assess risk without explicit failure labels in fixed action space. In contrast to these scenario-specific methods, another important technique is to use out-of-distribution (OOD) methods to detect failures. Fail-Detect~\cite{XuC2-RSS-25} formulate failure detection as an OOD problem to address the limitations of data collection. While prior works have made meaningful progress in failure detection, existing methods and benchmarks are evaluated on slow-speed manipulation scenarios and lack annotation of the critical intervention boundary\cite{mandlekar2022matters, XuC2-RSS-25}; therefore, we introduce a new dataset of fast-moving manipulation. 
\vspace{-1pt}

\subsection{OOD Methods for Failure Detection}

Out-of-distribution (OOD) detection provides a general framework for robotic failure detection. Distance-based methods such as Principal Component Analysis with K-means clustering (PCA-kmeans) embed observations via PCA, flagging observations far from the in-distribution clusters as failures~\cite{liu2025multi}. One-class methods like Random Network Distillation (RND) train a predictor to mimic a frozen random target network on successful trajectories, where a large mismatch at inference signals OOD behavior~\cite{he2024rediffuser}. Consistency Flow Matching (CFM) measures trajectory curvature in the observation-to-noise forward flow, where high curvature indicates that the input is OOD~\cite{yang2024consistency}. Second-order methods like Natural Posterior Network (NatPN) impose a Dirichlet prior on class probabilities to separate epistemic from aleatoric uncertainty, using high epistemic uncertainty as an indicator of OOD inputs~\cite{charpentier2022natural}. 

Building on these foundations, Fail-Detect~\cite{XuC2-RSS-25} combines scalar OOD scores with a time-varying conformal prediction band to guarantee runtime false positive rate control, introducing the log-probability in latent noise space (logpZO) score, which maps observations to latent noise via flow matching and uses the squared norm as the OOD signal. More recently, FIPER~\cite{romer2025failure} integrates an RND-based OOD score in the policy’s observation embedding space with an action-chunk entropy score over short sliding windows, triggering a failure alarm only when both exceed their conformal thresholds. As this is concurrent work, we do not include it as a baseline. However, existing OOD-based failure detection methods tend to over-confidently flag roll-outs as failure early in execution, motivating a more reliable failure detection scheme for fast-moving manipulation scenarios.

\section{PROBLEM FORMULATION}
\label{sec:problem}

We define proactive failure prediction for the lift-and-place task as a end-to-end time series prediction problem, which takes a sequence of visual observations and robot states as raw input. 
At each time step $t_i$, the system outputs a risk value $R_i \in \mathbb{R}$, representing the likelihood of a failure precursor. The observation $O_i = \{I_{t-m+1}, \ldots, I_i\}_\text{top,side}$ contains image frames $I$ from two camera ($C_\text{top},C_\text{side}$) within a temporal window of length $m$ (here we use $m=5$ empirically). The state $S_i$ at time $t_i$ is: 
\begin{equation}
\label{eq:states}
    S_i = \{(e_{i-m+1}, \theta_{i-m+1}, \dot{\theta}_{i-m+1}), \ldots, (e_i, \theta_i, \dot{\theta}_i)\},
\end{equation}
which contain the robot end-effector position ($e_i \in \mathbb{R}^3$), joint angles $\theta_i$, and joint velocities $\dot{\theta}_i$ for the past $m$ time steps.
In this paper, we study \emph{lift-and-place manipulation} in which a 6-DOF robot transports a target object $o$ on a flat carrier $c$ at maximum joint velocity. A failure occurs when the target object $o$ falls from the carrier $c$ while the robot is in motion. If the object reorients but remains on the carrier, the manipulation is considered successful. We assume at most one failure per trajectory, as a dropped object terminates the lift-and-place task.

\begin{figure}
    \centering
    \includegraphics[width=\linewidth]{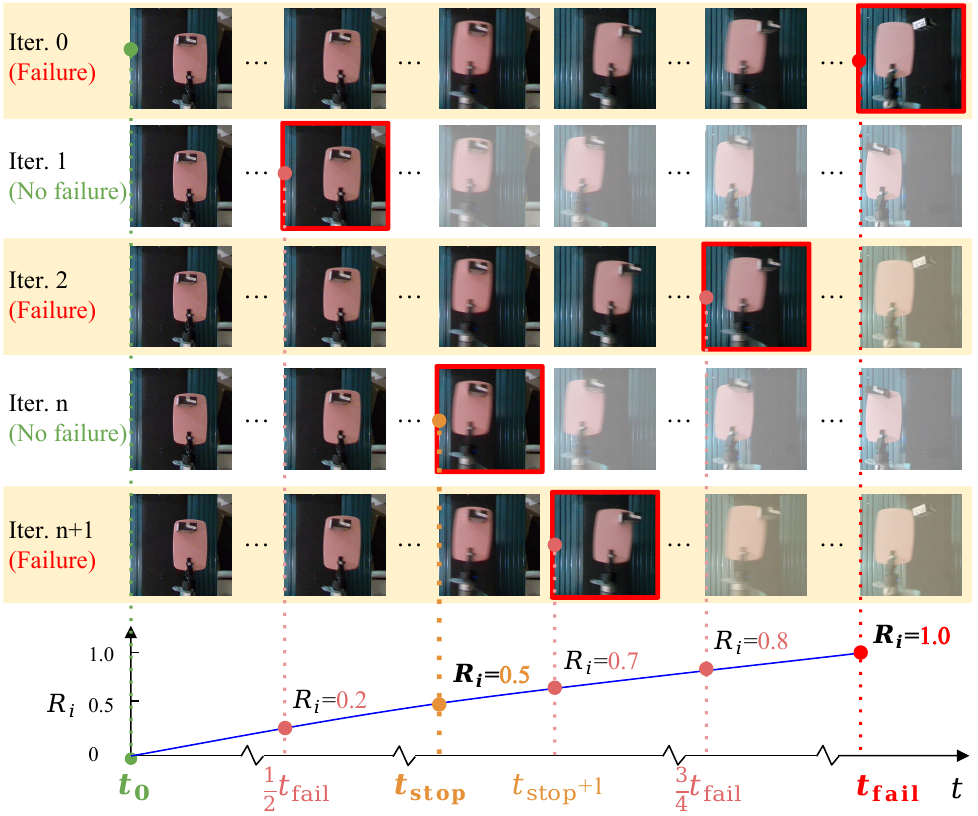}
    \caption{\textbf{Historical backtracking data collection pipeline.} The upper portion illustrates the iterative binary search process (across all iterations: Iter.~0,~1,~2, ...,~n,~n+1) for historically backtracking from the actual failure moment $t_\text{{fail}}$ (the black box drops from the pink panel), namely historical backtracking. In each iteration, we apply an e-stop (indicated by red border) and observe whether failure still occurs in the following time steps (indicated by the translucent color). The binary searching process will eventually converge to a critical time step $t_\text{stop}$, defined as the latest point at which applying an emergency stop would still prevent failure, such that the failure would not occur at $t_\text{stop}$ but would inevitably occur at the subsequent timestep $t_\text{stop} + 1$ . The curve in the bottom part represents the labeled ground truth value of the risk value, $R_i=0.5$ at the moment $t_\text{stop}$, $R_i=1.0$ at $t_\text{{fail}}$.}
    \label{fig:historical backtracking}
    \vspace{-20pt}
\end{figure}

\section{METHOD: \method}
\label{sec:method}

Our method, \method, consists of two input branches, as shown in Fig.~\ref{fig:Pipeline}. The observation branch extracts visual features from the multi-view video stream and the computed relative motion between the target object and the carrier. The state branch encodes robot states (from Eq.~\ref{eq:states}) along with predicted risk values from previous time steps. These features are processed to produce a continuous risk estimate. 

\subsection{Data Labeling: Historical Backtracking}
\label{sec:historical}

We design a data collection pipeline that assigns a risk value $R_i \in [0, 1]$ to each time step during manipulation, starting from the first time step $t_0$. We propose a \textit{historical backtracking method} for data collection in simulation, which identifies two key moments within a trajectory: (1) the last time step at which an emergency stop (e-stop) can prevent a failure, denoted as $t_\text{stop}$, and (2) the moment when the failure actually occurs, denoted as $t_\text{{fail}}$; for lift-and-place tasks, $t_\text{{fail}}$ is the moment when the object drops from the carrier. 

To determine $t_\text{stop}$ from historical backtracking, we use counterfactual reasoning~\cite{miller2025counterfactual} and iteratively search backward from $t_\text{{fail}}$ using a binary search strategy to find the earliest, safe intervention point. 
For each trajectory, we first identify $t_\text{fail}$ which is detected in simulation by checking whether the object's center of mass drops below that of the carrier, or manually annotated in the real world. We then iteratively narrow down $t_\text{stop}$ by: (1) replaying the trajectory and applying an e-stop at the midpoint of the current search interval, (2) observing whether failure still occurs, and (3) updating the interval accordingly (searching the earlier half if failure occurs, and the later half otherwise). As shown in Fig.~\ref{fig:historical backtracking}, in Iter.~1 we apply the e-stop at $\frac{1}{2}t_\text{fail}$ and observe no failure, so we search later; in Iter.~2 we apply it at $\frac{3}{4}t_\text{fail}$ and observe failure, so we search earlier. This process continues until we find $t^*$ such that an e-stop at $t^*$ prevents failure while one at $t^*+1$ does not, which we define as $t_\text{stop}$. Then, for other time points, we apply Hermite splines~\cite{erdogan2013spline} to interpolate the risk values, enabling the model to learn the temporal progression toward the boundary (see Sec.~\ref{sec:discussion}).

In the real world, historical backtracking requires physical resets of the same trajectory. 
Human annotators label three key moments: $t_0$ is the start of manipulation; $t_\text{{fail}}$ is the moment that the object starts falling off the carrier; and $t_{\text{stop}}$, which is determined via a physical binary search procedure. The rollout is repeatedly replayed with the emergency stop applied at different times, and $t_{\text{stop}}$ is the latest time at which applying the e-stop prevents failure. Because the e-stop is applied to the physical robot and its outcome directly observed, $t_{\text{stop}}$ is empirically verified. 




\subsection{Network Design}
\label{sec:network}

As illustrated in Fig.~\ref{fig:Pipeline}, our network adopts a two-branch architecture. 
The observation branch extracts visual features from multi-view video data and the relative motion computed from consecutive images. 
The state branch incorporates proprioceptive robot state variables together with predicted risk values from previous time steps.

\textbf{Image sequence feature.} 
Faster relative motion often correlates with higher risk. 
However, such motion may be ambiguous or unobservable from certain viewpoints, such as when the object moves along the camera's line of sight. To improve robustness, we use two cameras from distinct perspectives to capture complementary motion cues. 
Since robot manipulation is inherently dynamic, we conduct failure analysis over a temporal window of $m$ steps using RGB image sequences from both cameras. 
The image sequence $I$ is encoded by $E_1$ (here we use a CNN layer) to produce the visual features $P_1$.

\textbf{Relative motion feature.} At each time step $t_i$, we obtain the relative motion values $M$ between the target object $o$ and the carrier $c$ in each camera view ($C_\text{top}$ and $C_\text{side}$), represented by $M_\text{top}^i$ and $M_\text{side}^i$, from the image features. Specifically, $M_{\text{top}, \text{side}}^i = \{R_{o-c}^i, T_{o-c}^i\}_\text{top, side}$ includes an object-carrier relative rotation value ($R_{o-c}^i = \theta^i \in \mathbb{R}^1$) and an object-carrier relative translation value ($T_{o-c}^i = (X, Y)^i \in \mathbb{R}^2$). In each camera view $C$, the object-carrier relative rotation $R^i$ is computed as the difference between the object-object relative rotation ($R_{o-o}^i$) and the carrier-carrier relative rotation ($R_{c-c}^i$), where the same decomposition holds for the object-carrier relative translation:
\begin{equation}
    R_{o-c}^i = R_{o-o}^i - R_{c-c}^i, \quad
    T_{o-c}^i = T_{o-o}^i - T_{c-c}^i
    \label{eq:o-c relative motion}
\end{equation}
To obtain the object-object and carrier-carrier relative motions in Eq.~\ref{eq:o-c relative motion}, we utilize 2D iterative closest point (ICP)~\cite{besl1992method, Yang_2013_ICCV}, which estimates the relative displacement between the object's (or carrier's) position at the current time step $t_i$ and its position at the initial time step $t_0$:
\begin{equation}
    (R,T)_{o-o,c-c}^i = ICP(V^0, V^i)
    \label{eq:o-o relative motion}
\end{equation}
where $V^0$ and $V^i$ denote the geometric contours of the target object or carrier at the initial and current time steps, respectively. Concretely, we extract the contours $\{V_o^{i}, V_c^{i}\}_\text{top, side}$ in both camera views at each time step $i$: in real-world settings, we leverage a pre-trained SAM2 model~\cite{ICLR2025_45c1f6a8} to segment and obtain these contours, while in simulation we directly use the projected object shape as its contour. We apply 2D ICP to these contours to compute the motion values $(R,T)_{o-o,c-c}^i$, and its accuracy is further validated in the supplementary material. The relative motion makes it readily applicable to other non-prehensile manipulation tasks, since SAM2 can be used to segment arbitrary objects, instead of relying on prior knowledge of object geometry.
As shown in Fig.~\ref{fig:Pipeline}, the relative motion $M$ passes through an encoder $E_2$ (here we use a MLP layer) to produce the feature embeddings $P_2$.

\textbf{Robot state feature.} In our second branch, we process the robot state $S$ as an additional feature, which includes the end-effector position, joint angles, and joint speeds, as well as the predicted risk values from the past $m-1$ steps within the current sampling window. The first three values capture the physical state of the robot and, since they are correlated with the image observations across different camera views, this feature can be shared with the visual observation features. The past risk values encode the temporal trend of the manipulation risk, leading to more robust inference (see Sec.~\ref{sec:ablation}). Although derivable from joint angles via forward kinematics, we provide the end-effector position explicitly to ease training. These states are passed through an encoder $E_3$ (we use a MLP layer) to produce the feature embedding $P_3$.

Finally, $P_1$, $P_2$, and $P_3$ are concatenated and passed through an encoder $E_4$ (here we use a ResNet module~\cite{he2016deep}) to produce the final continuous risk value $R_i \in [0,1]$. In a training batch, the samples with high risk value and the samples with low risk value are evenly mixed for balanced training. Also, the past risk is disturbed randomly for data augmentation purpose, similar to teacher forcing~\cite{lamb2016professor}. Details are in the supplementary website.


\begin{table*}[t]
    \centering
	\resizebox{\textwidth}{!}{
    \begin{tabular}{cccccccccc}
    \toprule
    \multicolumn{2}{c}{\multirow{2}{*}{Methods / Variants}} & \multicolumn{4}{c}{Simulation} & \multicolumn{4}{c}{Real-world} \\
    \cmidrule(lr){3-6} \cmidrule(lr){7-10}
    \multicolumn{2}{c}{} & Precision $\uparrow$ & Accuracy $\uparrow$ & Recall $\uparrow$ & $\eta \uparrow$ & Precision $\uparrow$ & Accuracy $\uparrow$ & Recall $\uparrow$ & $\eta \uparrow$ \\
    \midrule
    \multirow{10}{*}{\shortstack{Fail-Detect~\cite{XuC2-RSS-25}\\(ResNet18)}} & (CFM~\cite{yang2024consistency}, FM) & 0.389 & 0.389 & \textbf{1.000} & 0.179 & 0.478 & 0.478 & \textbf{1.000} & 0.297\\
    & (CFM~\cite{yang2024consistency}, DP) & 0.386 & 0.386 & \textbf{1.000} & 0.202 & 0.500 & 0.500 & \textbf{1.000} & 0.303\\
    & (NatPN~\cite{charpentier2022natural}, FM) & 0.374 & 0.374 & \textbf{1.000} & 0.198 & 0.493 & 0.493 & \textbf{1.000} & 0.275\\
    & (NatPN~\cite{charpentier2022natural}, DP) & 0.384 & 0.384 & \textbf{1.000} & 0.185 & 0.493 & 0.493 & \textbf{1.000} & 0.275\\
    & (PCA-kmeans~\cite{liu2025multi}, FM) & 0.389 & 0.389 & \textbf{1.000} & 0.104 & 0.493 & 0.493 & \textbf{1.000} & 0.226\\
    & (PCA-kmeans~\cite{liu2025multi}, DP) & 0.388 & 0.388 & \textbf{1.000} & 0.082 & 0.489 & 0.489 & \textbf{1.000} & 0.221\\
    & (RND~\cite{he2024rediffuser}, FM) & 0.369 & 0.369 & \textbf{1.000} & 0.267 & 0.504 & 0.504 & \textbf{1.000} & 0.289\\
    & (RND~\cite{he2024rediffuser}, DP) & 0.384 & 0.384 & \textbf{1.000} & 0.170 & 0.507 & 0.507 & \textbf{1.000} & 0.302\\
    & (logpZO~\cite{XuC2-RSS-25}, FM) & 0.389 & 0.389 & \textbf{1.000} & 0.118 & 0.504 & 0.504 & \textbf{1.000} & 0.195\\
    & (logpZO~\cite{XuC2-RSS-25}, DP) & 0.389 & 0.389 & \textbf{1.000} & 0.083 & 0.496 & 0.496 & \textbf{1.000} & 0.219\\
    \midrule
    \multicolumn{2}{c}{\textbf{\method (ours)} (ResNet18)} & 0.991 & 0.995 & 0.995 & 0.933 & \textbf{0.944} & \textbf{0.963} & 0.985 & \textbf{0.932} \\
    \multicolumn{2}{c}{\textbf{\method (ours)} (ResNet50)} & 0.967 & 0.986 & 0.995 & \textbf{0.964} & 0.916 & 0.948 & 0.984 & 0.910\\
    \multicolumn{2}{c}{\textbf{\method (ours)} (ResNet152)} & \textbf{0.995} & 0.997 & 0.995 & 0.923 & 0.900 & 0.933 & 0.978 & 0.925\\
    \bottomrule
    \end{tabular}
    }
    \caption{Performance on simulation and real-world data. For simulation, the model is both trained and tested on simulation data. For real-world, the model is trained and tested on real-world data, with results reported as the average over 10-fold cross-validation across 10 experiments. The baseline method has multiple variants, each evaluated with different policy generation methods (FM for flow-matching~\cite{lipman2023flow}, DP for diffusion policy~\cite{chi2025diffusion}). The best results are highlighted in bold. Our \method consistently and largely outperforms all baseline variants. Although we do not achieve 1.000 recall, our high precision and high recall together indicate a stronger ability to distinguish failures from non-failure modes. In contrast, the baseline's low precision with high recall suggests a biased output that defaults to predicting failure. For more details, see Sec.~\ref{sec:sim result}.}
    \label{tab:experiment_sim}
    \vspace{-10pt}
\end{table*}

\begin{table}[t]
    \centering
	\resizebox{0.5\textwidth}{!}{
    \begin{tabular}{ccccc}
    \toprule
    Modules & Precision $\uparrow$ & Accuracy $\uparrow$ & Recall $\uparrow$ & $\eta \uparrow$ \\
    \midrule
    \textbf{Complete} & \textbf{0.995} & \textbf{0.997} & 0.995 & 0.923\\
    \midrule
    Classification loss only & 0.574 & 0.850 & \textbf{1.000} & \textbf{0.990}\\
    Regression loss only & 0.980 & 0.991 & 0.995 & 0.855\\
    \midrule
    Without images & 0.973 & 0.988 & 0.995 & 0.904\\
    Without relative motion & 0.812 & 0.870 & 0.816 & 0.596\\
    \midrule
    Without past risk & 0.314 & 0.315 & \textbf{1.000} & 0.832\\
    Without end-effector position & 0.958 & 0.983 & 0.995 & 0.910\\
    Without joint angles & 0.927 & 0.972 & \textbf{1.000} & 0.860\\
    Without joint speeds & 0.384 & 0.417 & 0.995 & 0.839\\
    \bottomrule
    \end{tabular}
    }
    \caption{Ablation results of \method trained on simulation data. We first decouple the loss terms by disabling classification or regression, then isolate the relative motion from the observation features, and then partially remove the robot state. The results indicate that the past risk values and the relative motion are the most critical components. Furthermore, both losses contribute to balancing precision, accuracy, recall, and intervention timeliness $\eta$.}
    \label{tab:ablation}
    \vspace{-15pt}
\end{table}


For a training batch with $N$ samples, the total training loss $\mathcal{L}_\text{Total}$ is defined as a weighted combination of a classification loss $\mathcal{L}_\text{Cls}$ and a regression loss $\mathcal{L}_\text{Reg}$:
\begin{equation} 
    \mathcal{L}_\text{Total} = \alpha \mathcal{L}_\text{Cls} + (1-\alpha) \mathcal{L}_\text{Reg},
    \label{eq:all}
\end{equation}
where $\alpha$ controls the relative contribution of each term. The regression loss uses mean squared error (MSE) to measure the discrepancy between the predicted risk values $R_i^\text{pred}$ and their ground truth counterparts $R_i^\text{gt}$, enabling the model to learn the temporal evolution of risk,
\begin{equation}
    \mathcal{L}_\text{Reg} = \frac{1}{N}\sum_{t=1}^{N} (R_i^\text{pred} - R_i^\text{gt})^2.
    \label{eq:reg}
\end{equation}
Since the learning objective is also to identify the safety boundary for triggering an emergency stop, each training sample is assigned a binary label $y^i$ indicating whether it corresponds to a high-risk state (after $t_\text{stop}$) or a low-risk state (before $t_\text{stop}$). The classification loss is formulated as binary cross-entropy,
\begin{equation}
    \mathcal{L}_\text{Cls} = -\frac{1}{N}\sum_{i=1}^{N} \left[ y_i \log(\hat{y}_i) + (1-y_i)\log(1-\hat{y}_i) \right],
    \label{eq:cls}
\end{equation}
which strengthens the model's ability to discriminate between safe and unsafe conditions, and consequently will improve its identification of $t_\text{stop}$ and leading to a higher intervention timeliness ratio $\eta$ (see Sec.~\ref{sec:assessment} and Sec.~\ref{sec:ablation}).

\subsection{Assessment Metrics}
\label{sec:assessment}

\begin{figure*}[t]
    \centering
    \includegraphics[width=\linewidth]{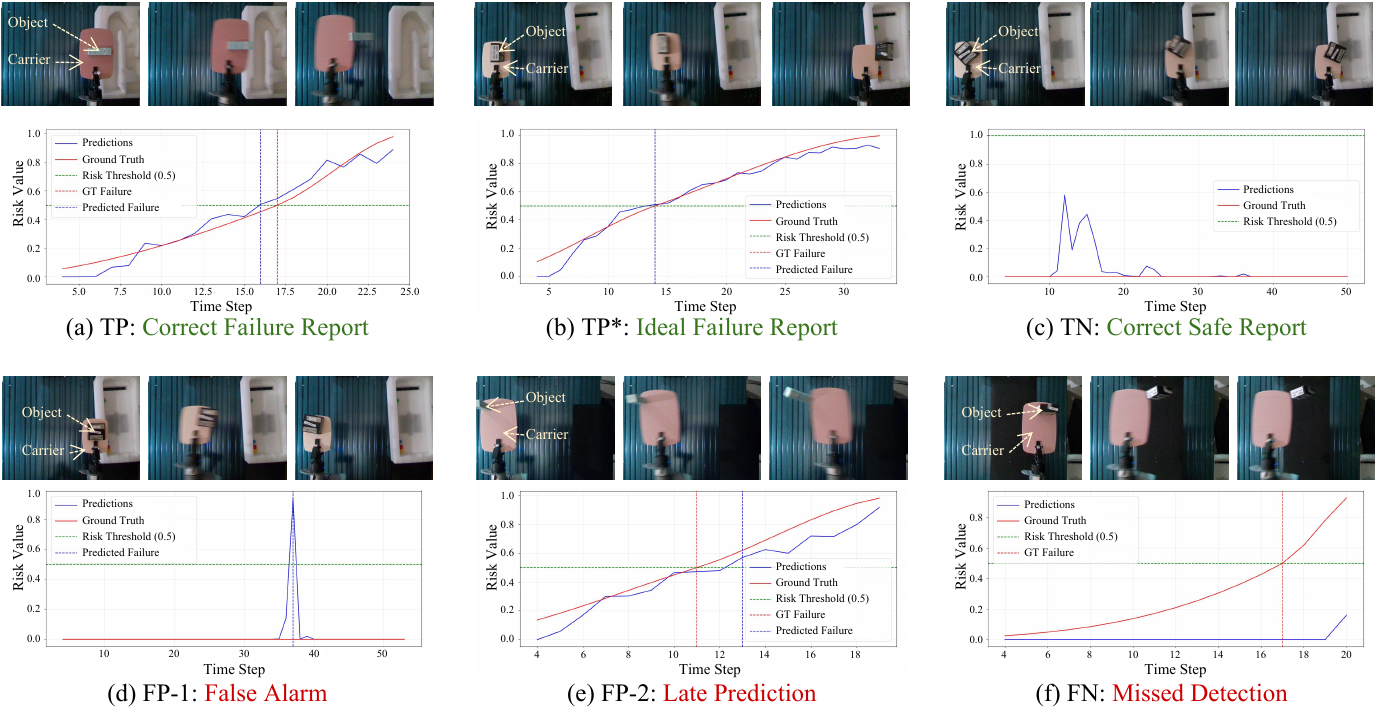}
    \caption{Visualization of all evaluation results on real-world test cases, with the predicted risk value curve (solid red line) vs the ground truth risk value curve (solid blue line) during evaluation. The dotted red line refers to the moment of $t_\text{stop}$, while the dotted blue line is $t_\text{stop}^\text{pred}$. The threshold of risk value ($R_i=0.5$) is indicated by dotted green line. TP* is a perfect case that demonstrates $\eta = 1.0$. See Sec.~\ref{sec:assessment} for details.}
    \label{fig:visualization}
    \vspace{-15pt}
\end{figure*}

We evaluate using four metrics, precision, accuracy, recall, and intervention timeliness ratio. We calculate precision, accuracy, and recall by defining 5 cases. A true positive (TP) occurs when the model correctly predicts a failure before the latest intervention point $t_\text{stop}$, leaving sufficient time for effective action. A true negative (TN) corresponds to correctly identifying a successful rollout. A false negative (FN) occurs when the model misses a failure. False positives (FP) include two distinct cases: FP-1, where the model incorrectly reports a failure for a successful rollout, and FP-2, where the model detects a real failure only after $t_\text{stop}$, rendering the prediction too late. Examples are in Fig.~\ref{fig:visualization}.


\begin{figure}[t]
    \centering
    \includegraphics[width=\linewidth]{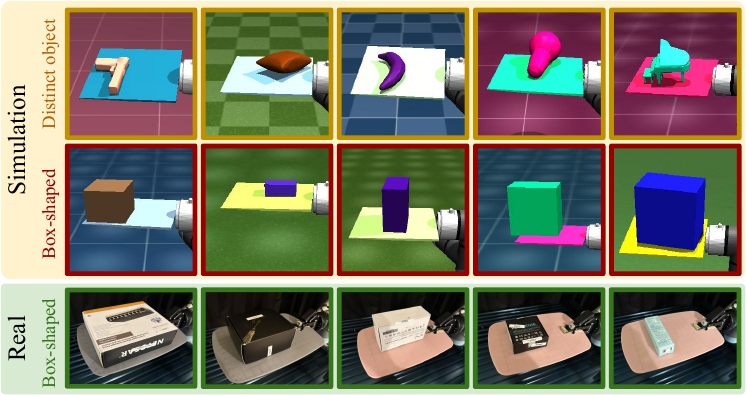}
    \caption{Visualization of the objects and carriers used in our dataset.}
    \label{fig:dataset}
    \vspace{-20pt}
\end{figure}

In addition, to discourage overly conservative predictions, we introduce the intervention timeliness ratio $\eta$. This represents the percentage of time steps completed at the moment when the failure is correctly predicted: 
\begin{equation}
    \eta = t_\text{stop}^\text{pred}/t_\text{stop},
    \label{eq:execution_rate}
\end{equation}
where the optimal $\eta$ is 1.0, indicating that the model predicts the failure at the last possible moment before the intervention deadline, thus maximizing task execution progress before triggering an emergency compensation, as shown in Fig.~\ref{fig:visualization}. Given the fast-moving task, a higher $\eta$ is desirable, as it implies that the intervention minimizes the overall cycle time $T$. 
For real-world examples, see Sec.~\ref{sec:deployment}.

\section{EXPERIMENTS}
\label{sec:experiments}

\subsection{Data and Experiment Setup}

We collect two datasets to evaluate our method: one in simulation using MuJoCo~\cite{todorov2012mujoco} and one in the real world. 
Both use a UR5 robot arm with a flat tray carrier holding an object placed on top. The carrier is attached via a Robotiq 2F-85 parallel-jaw gripper in the real world, and rigidly fixed to the end effector through a defined joint in simulation. In the real world, two Intel RealSense D435 cameras provide top-down and side views in 30fps.

As discussed in Sec.~\ref{sec:historical}, collecting real-world data is labor-intensive, so we collect data in simulation to facilitate large-scale verification. The \textbf{simulated} dataset has two subsets. The first focuses on box-shaped objects to mimic typical industrial handling scenarios~\cite{sarawgi2026binpacking} and contains 539 simulation trajectories. The second contains 5,236 simulation trajectories with diverse daily objects adopted from Lum~et~al.~\cite{pmlr-v270-lum25b}; training and test trajectories are drawn from this set to evaluate performance across a wide range of geometries. In addition, we collect 254 \textbf{real-world} trajectories using five types of box-shaped objects. Two boxes include packaged goods with shifting contents, which introduces more variability. Experiments are conducted separately on simulation and real-world data. For each dataset, we split the collected trajectories into training, validation, and test sets with an 8:1:1 ratio (see Fig.~\ref{fig:dataset} for examples). See supplementary website for detailed description of the dataset.

Each trajectory corresponds to a rapid lift-and-place task in which the robot dynamically moves the carrier (with an object on it) within the workspace, operating at its \textbf{maximum} joint velocity.
To simulate challenging industrial conditions and increase trajectory diversity, we randomize object and physics parameters during data collection; details are provided in the supplementary website.

\subsection{Model Evaluation on Simulated Data}
\label{sec:sim result}

We train and evaluate both the baseline method~\cite{XuC2-RSS-25} and our model on the 5,775 simulated trajectories. All methods are trained for 200 epochs, with results reported in Tab.~\ref{tab:experiment_sim}. As described in Sec.~\ref{sec:assessment}, the four metrics assess performance from complementary perspectives. Compared to the OOD baseline, \method significantly improves overall performance. For example, precision improves by over 155\% (from 0.389 to 0.995) and accuracy by over 156\% (from 0.389 to 0.997), reducing respective error rates by more than 99\%, demonstrating that \method substantially reduces false positives while correctly identifying true failure cases.

Although the baseline achieves a recall of 1.000 on our dataset, this comes with very low precision and accuracy. Notably, its precision and recall are identical, indicating that the baseline predicts every trajectory as failure. Such behavior is overly conservative and fails to distinguish between true positives (TP) and true negatives (TN). In contrast, \method achieves both high precision and high recall, showing that it can effectively discriminate between failure and non-failure trajectories.
In addition, \method improves the execution-aware metric $\eta$ from 0.198 to 0.964, indicating substantially more accurate localization of the critical stopping time $t_{\text{stop}}$.

\subsection{Model Evaluation on Real-world Data}
\label{sec:real result}


We train and evaluate our model on 254 real-world trajectories. 
To ensure robustness given the limited data size, we perform a ten-fold cross-validation by constructing three different splits following the same ratio and report the average performance in Tab.~\ref{tab:experiment_sim}.




\subsection{Component Analysis}
\label{sec:ablation}

To evaluate the contribution of each module in our network design, we perform ablation studies using the ResNet152 backbone on the simulation dataset from three perspectives: (1) loss design, (2) observation inputs, and (3) robot state inputs. The results are summarized in Tab.~\ref{tab:ablation}.

\textbf{Loss design}. We analyze the individual effects of the classification and regression losses. As shown in Tab.~\ref{tab:ablation}, using only classification loss results in lower precision and accuracy but a high intervention timeliness ratio, whereas using only regression loss yields high precision and accuracy but a lower intervention timeliness ratio. These findings are consistent with our analysis in Sec.~\ref{sec:network}, where the classification objective primarily improves failure discrimination and enhances localization of $t_{\text{stop}}$, while the regression objective temporal evolution of the risk value.

\textbf{Observation inputs}. We evaluate the contributions of image sequences and relative motion features. As reported in Tab.~\ref{tab:ablation}, using relative motion or image sequences alone does not achieve optimal performance. Combining both modalities improves performance, indicating that visual and motion cues provide complementary information, with relative motion playing a more dominant role.

\textbf{Robot state inputs}. 
We train the model by excluding one robot state at a time to analyze the contribution of each input component. Tab.~\ref{tab:ablation} shows that past risk values and joint speed data contribute significantly to the overall performance.

\subsection{Visualization}
\label{sec:visualization}

In Fig.~\ref{fig:visualization}, we visualize representative prediction results and simultaneously illustrate the five cases used to calculate precision and recall. These results are obtained under Sim\&Real training and evaluation using ResNet-18.
In the TN case, our model correctly predicts low risk values throughout the manipulation.
In the FP case, the model either incorrectly predicts a high risk value when no failure occurs, or it predicts $t_\text{stop}^\text{pred}$ after the ground-truth $t_\text{stop}$.
In the TP case, the model correctly predicts $t_\text{stop}^\text{pred}$ before the ground-truth $t_\text{stop}$.
In the FN case, the model fails to predict any high risk values, even though the ground truth indicates a failure.
Additionally, the TP* case represents an optimal prediction, where $t_\text{stop}^\text{pred} = t_\text{stop}$ and $\eta = 1.0$.

\subsection{Real-world Deployment}
\label{sec:deployment}

To validate real-world applicability, we deployed \method on the lift-and-place task (Sec.~\ref{sec:problem}) and used its predicted risk to proactively prevent failures.

\begin{figure}
    \centering
    \includegraphics[width=\linewidth]{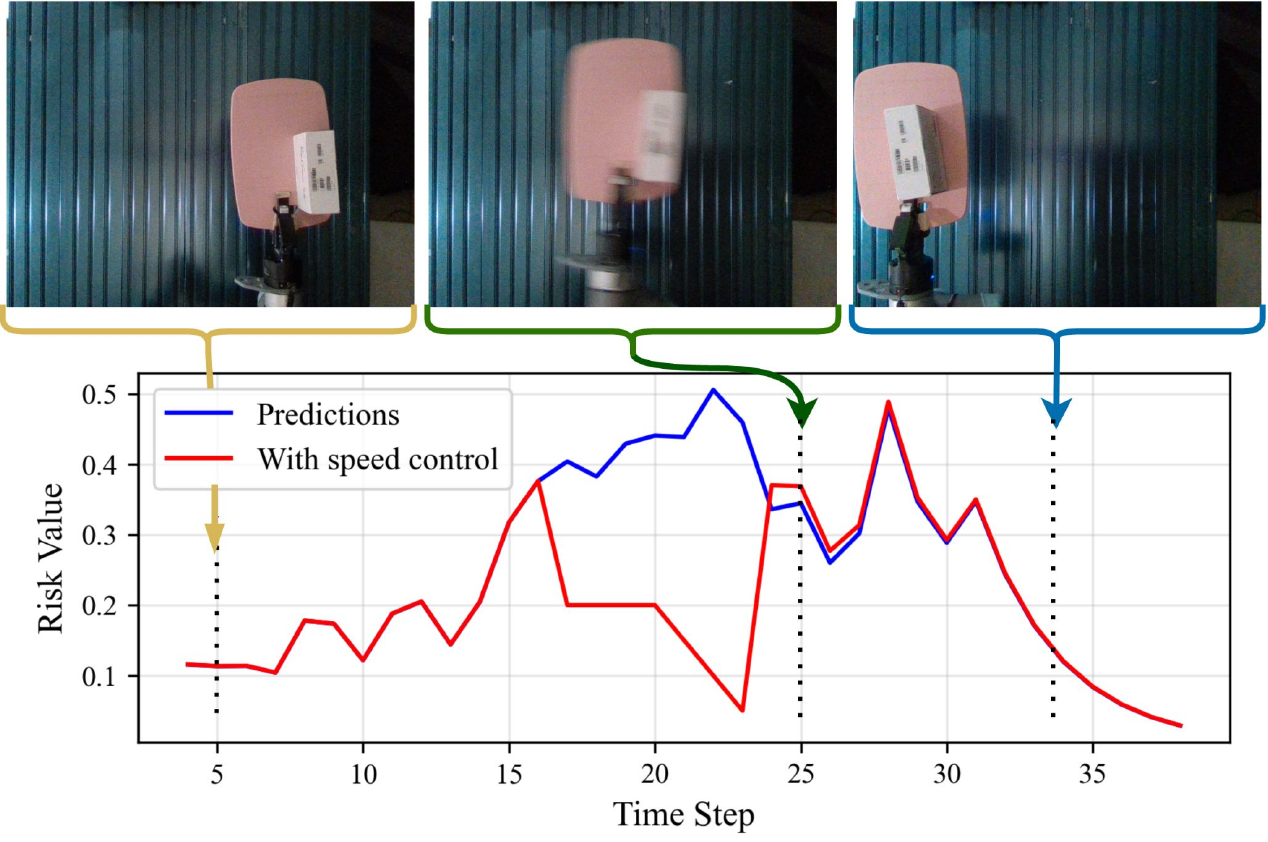}
    \caption{Trajectory showing a real-world deployment scenario where joint speed is set based on the predicted risk value which in turn mitigates subsequent risk after time step 20.}
    \label{fig:Deploy_placeholder}
    \vspace{-20pt}
\end{figure}

In a lift-and-place manipulation that potentially fails before a total cycle time of $T$, our goal is to intervene at an early stage by modulating the joint speed to prevent the failure, incurring an additional cycle time $\Delta T$. For example, the original rollout failed around \SI{0.86}{\second}, although it was expected to finish in \SI{1.2}{\second}. With our speed control strategy, the robot instead completes the task successfully in \SI{1.3}{\second}, incurring only a \SI{0.1}{\second} increase in cycle time. By contrast, slowing down from the very beginning would require at least \SI{1.8}{\second} to avoid failure. We use the predicted risk to dynamically modulate joint velocities. At risk 0.0, the robot operates at maximum speed; as risk increases, speed decreases proportionally, reaching the minimum at risk 0.5. Speeds in between are determined by linear interpolation, as shown in Fig.~\ref{fig:Deploy_placeholder}.
This risk-aware velocity scaling enables the system to respond adaptively to potential failure states, slowing down cycle in higher-risk scenarios to improve task robustness while maintaining efficiency in low-risk conditions.

\subsection{Discussion}
\label{sec:discussion}

\textbf{Limitations.} \method has several limitations that present opportunities for future work. First, we study a single object on a carrier. Extending \method to multi-object scenarios would require reasoning about inter-object interactions and determining which relative motions are safety-critical. 
Second, while our method predicts when intervention is needed, it does not address a general \emph{recovery strategies}, which are application-specific and may require domain knowledge or specialized hardware.
Third, our relative motion estimation relies on 2D ICP over projected contours, which may degrade under occlusion, poor lighting, or visual ambiguity; depth sensing with 3D ICP could provide more robustness at increased computational cost. 
Finally, our historical backtracking data collection is efficient in simulation but is labor-intensive in the real-world due to repeated rollouts and resets. 

\textbf{Can we model the system dynamics analytically and predict failure from initial conditions and planned trajectories?} The contact dynamics between objects and carriers involve complex, nonlinear interactions governed by friction, surface geometry, and material properties that are not reliably identifiable online and are difficult to model accurately. This motivates a learning-based approach that infers failure risk directly from observed data. 

\textbf{Why use interpolated values for the risk curve?} Our primary goal is to accurately identify the critical boundary $t_\text{stop}$. Intermediate risk values act as supervisory signals that help the model learn the temporal progression toward this boundary, approximating the continuous degradation of recoverability as destabilizing factors (momentum, contact instability) accumulate, and the safety margin diminishes smoothly. Smooth continuous labels provide better learning signals than discrete jumps. Although exact intermediate values are not physically measured, they provide a relative ordering of risk that helps the model distinguish early-stage stable conditions from late-stage critical conditions. The classification loss enforces learning of the binary boundary ($R_i$ = 0.5 at $t_\text{stop}$), while the regression loss encourages temporal consistency. In preliminary experiments, spline interpolation outperformed step and exponential labeling in striking best balance in intervention timeliness and precision (see supplementary website for details).

\textbf{Why use a fixed threshold?} While failure patterns vary across rollouts (e.g., different object dynamics), our model learns to map diverse observable patterns to a calibrated risk scale where 0.5 consistently represents the $t_\text{stop}$ boundary. The network implicitly learns: ``when I observe motion pattern X, how close am I to $t_\text{stop}$?''
A fixed threshold enables consistent interpretation: operators know that $R_i > 0.5$ means past the ``point of no return,'' simplifying deployment across object types and speeds without per-scenario tuning.

\section{CONCLUSION}
\label{sec:conclusion}

In this paper, we investigate how proactive failure prediction can balance cycle time and risk management in non-prehensile lift-and-place material handling tasks. We propose \method, which extracts relative motion from images to capture scene dynamics and predict risk in advance. We demonstrate that \method achieves an effective guarantee of cycle time and manipulation reliability by adapting execution speed based on risk assessment. Our experimental investigation focused on object transport using a part carrier. However, we believe that the methodology presented in this paper will advance proactive robotic failure prediction for a wide variety of high-speed material handling applications.






\section*{ACKNOWLEDGMENT}

{\small 
We acknowledge support from the National Artificial Intelligence Research Resource (NAIRR) Pilot program. We thank Jeremy Morgan, Jason Chen, Siheng Zhao, Minjune Hwang, and Zhehui Huang for their helpful technical advice and discussions. 
}


\bibliographystyle{IEEEtran}
\bibliography{references}

@String { icra    = {IEEE International Conference on Robotics and Automation (ICRA)} }

@String { ieeera  = {IEEE Robotics and Automation Letters (RA-L)} }

@String { ijrr    = {International Journal of Robotics Research (IJRR)} }

@String { iros    = {IEEE/RSJ International Conference on Intelligent Robots and Systems (IROS)} }

@String { rss     = {Robotics: Science and Systems (RSS)} }

@String { tase    = {IEEE Transactions on Automation Science and Engineering}}

@String { neurips = {Neural Information Processing Systems (NeurIPS)} }

@String { icml    = {International Conference on Machine Learning (ICML)} }

@String { iclr    = {International Conference on Learning Representations (ICLR)} }

@String { corl    = {Conference on Robot Learning (CoRL)} }

@String { hri     = {International Conference on Human-Robot Interaction (HRI)} }

@String { cvpr    = {IEEE Conference on Computer Vision and Pattern Recognition (CVPR)} }

@String { iccv    = {IEEE/CVF International Conference on Computer Vision (ICCV)} }

@inproceedings{sarawgi2026binpacking,
  title={{Expanding Picking Actions for Time-Efficient Online 3D Bin Packing}},
  author={Nikita Sarawgi and Omey M. Manyar and Fan Wang and Thinh H. Nguyen and Daniel Seita and Satyandra K. Gupta},
  booktitle=icra,
  Year={2026}
}

@inproceedings{lemasurier2024reactive,
  title={Reactive or proactive? how robots should explain failures},
  author={LeMasurier, Gregory and Gautam, Alvika and Han, Zhao and Crandall, Jacob W and Yanco, Holly A},
  booktitle=hri,
  year={2024}
}

@inproceedings{ma2024multimodal,
  title={Multimodal Failure Prediction for Vision-based Manipulation Tasks with Camera Faults},
  author={Ma, Yuliang and Liu, Jingyi and Mamaev, Ilshat and Morozov, Andrey},
  booktitle=iros,
  year={2024}
}

@inproceedings{pramanick2024multimodal,
  title={Multimodal Coherent Explanation Generation of Robot Failures},
  author={Pramanick, Pradip and Rossi, Silvia},
  booktitle=iros,
  year={2024}
}

@inproceedings{zhou2025code,
  title={Code-as-monitor: Constraint-aware visual programming for reactive and proactive robotic failure detection},
  author={Zhou, Enshen and Su, Qi and Chi, Cheng and Zhang, Zhizheng and Wang, Zhongyuan and Huang, Tiejun and Sheng, Lu and Wang, He},
  booktitle=cvpr,
  year={2025}
}

@ARTICLE{Inceoglu2023Multimodal,
  author={Inceoglu, Arda and Aksoy, Eren Erdal and Sariel, Sanem},
  journal=ieeera, 
  title={Multimodal Detection and Classification of Robot Manipulation Failures}, 
  year={2024}
}

@inproceedings{liu2023reflect,
  title={REFLECT: Summarizing Robot Experiences for Failure Explanation and Correction},
  author={Liu, Zeyi and Bahety, Arpit and Song, Shuran},
  booktitle=corl,
  year={2023}
}

@inproceedings{kang2024using,
  title={Using Large Language Models to Generate and Apply Contingency Handling Procedures in Collaborative Assembly Applications},
  author={Kang, Jeon Ho and Dhanaraj, Neel and Wadaskar, Siddhant and Gupta, Satyandra K},
  booktitle=icra,
  year={2024}
}

@inproceedings{ICLR2025_70a06501,
 author = {Duan, Jiafei and Pumacay, Wilbert and Kumar, Nishanth and Wang, Yi Ru and Tian, Shulin and Yuan, Wentao and Krishna, Ranjay and Fox, Dieter and Mandlekar, Ajay and Guo, Yijie},
 booktitle = iclr,
 title = {AHA: A Vision-Language-Model for Detecting and Reasoning Over Failures in Robotic Manipulation},
 year = {2025}
}

@article{huang2021task,
  title={Task failure prediction for wafer-handling robotic arms by using various machine learning algorithms},
  author={Huang, Ping Wun and Chung, Kuan-Jung},
  journal={Measurement and Control},
  year={2021}
}

@inproceedings{
    sagar2026uncovering,
    title={Uncovering Robot Vulnerabilities through Semantic Potential Fields},
    author={Som Sagar and Jiafei Duan and Sreevishakh Vasudevan and Yifan Zhou and Heni Ben Amor and Dieter Fox and Ransalu Senanayake},
    booktitle=iclr,
    year={2026}
}

@inproceedings{ICLR2025_45c1f6a8,
 author = {Ravi, Nikhila and Gabeur, Valentin and Hu, Yuan-Ting and Hu, Ronghang and Ryali, Chaitanya and Ma, Tengyu and Khedr, Haitham and R\"{a}dle, Roman and Rolland, Chloe and Gustafson, Laura and Mintun, Eric and Pan, Junting and Alwala, Kalyan Vasudev and Carion, Nicolas and Wu, Chao-Yuan and Girshick, Ross and Dollar, Piotr and Feichtenhofer, Christoph},
 booktitle = iclr,
 title = {SAM 2: Segment Anything in Images and Videos},
 year = {2025}
}

@inproceedings{todorov2012mujoco,
  title={Mujoco: A physics engine for model-based control},
  author={Todorov, Emanuel and Erez, Tom and Tassa, Yuval},
  booktitle=iros,
  year={2012}
}

@InProceedings{pmlr-v270-lum25b,
  title = 	 {Get a Grip: Multi-Finger Grasp Evaluation at Scale Enables Robust Sim-to-Real Transfer},
  author =       {Lum, Tyler Ga Wei and Li, Albert H. and Culbertson, Preston and Srinivasan, Krishnan and Ames, Aaron and Schwager, Mac and Bohg, Jeannette},
  booktitle =   corl,
  year = 	 {2025}
}

@article{erdogan2013spline,
  title={Spline interpolation techniques},
  author={Erdogan, KAYA},
  journal={Journal of Technical Science and Technologies},
  year={2013}
}

@inproceedings{thoduka2024multimodal,
  title={A Multimodal Handover Failure Detection Dataset and Baselines},
  author={Thoduka, Santosh and Hochgeschwender, Nico and Gall, Juergen and Pl{\"o}ger, Paul G},
  booktitle=icra,
  year={2024}
}

@INPROCEEDINGS{sinha2024real, 
    AUTHOR    = {Rohan Sinha AND Amine Elhafsi AND Christopher Agia AND Matt Foutter AND Edward Schmerling AND Marco Pavone}, 
    TITLE     = {{Real-Time Anomaly Detection and Reactive Planning with Large Language Models}}, 
    BOOKTITLE = rss, 
    YEAR      = {2024}
}

@INPROCEEDINGS{XuC2-RSS-25, 
    AUTHOR    = {Chen Xu AND Tony Khuong Nguyen AND Emma Dixon AND Christopher Rodriguez AND Patrick Miller AND Robert Lee AND Paarth Shah AND Rares Andrei Ambrus AND Haruki Nishimura AND Masha Itkina}, 
    TITLE     = {{Can We Detect Failures Without Failure Data? Uncertainty-Aware Runtime Failure Detection for Imitation Learning Policies}}, 
    BOOKTITLE = rss, 
    YEAR      = {2025}
}

@article{yang2024consistency,
  title={Consistency flow matching: Defining straight flows with velocity consistency},
  author={Yang, Ling and Zhang, Zixiang and Zhang, Zhilong and Liu, Xingchao and Xu, Minkai and Zhang, Wentao and Meng, Chenlin and Ermon, Stefano and Cui, Bin},
  journal={arXiv preprint arXiv:2407.02398},
  year={2024}
}

@inproceedings{
    charpentier2022natural,
    title={Natural Posterior Network: Deep Bayesian Predictive Uncertainty for Exponential Family Distributions},
    author={Bertrand Charpentier and Oliver Borchert and Daniel Z{\"u}gner and Simon Geisler and Stephan G{\"u}nnemann},
    booktitle=iclr,
    year={2022}
}

@inproceedings{liu2025multi,
  title={Multi-Task Interactive Robot Fleet Learning with Visual World Models},
  author={Liu, Huihan and Zhang, Yu and Betala, Vaarij and Zhang, Evan and Liu, James and Ding, Crystal and Zhu, Yuke},
  booktitle=corl,
  year={2025}
}

@inproceedings{he2024rediffuser,
  title={ReDiffuser: reliable decision-making using a diffuser with confidence estimation},
  author={He, Nantian and Li, Shaohui and Li, Zhi and Liu, Yu and He, You},
  booktitle=icml,
  year={2024}
}

@inproceedings{
    romer2025failure,
    title={Failure Prediction at Runtime for Generative Robot Policies},
    author={Ralf R{\"o}mer and Adrian Kobras and Luca Worbis and Angela P. Schoellig},
    booktitle=neurips,
    year={2025}
}

@inproceedings{parashar2025failure,
  title={Failure Prediction from Limited Hardware Demonstrations},
  author={Parashar, Anjali and Garg, Kunal and Zhang, Joseph and Fan, Chuchu},
  booktitle={2025 61st Allerton Conference on Communication, Control, and Computing Proceedings},
  year={2025}
}

@article{lee2025grasp,
  title={Grasp failure constraints for fast and reliable pick-and-place using multi-suction-cup grippers},
  author={Lee, Jee-eun and Sun, Robert and Bylard, Andrew and Sentis, Luis},
  journal=tase,
  year={2025}
}

@inproceedings{zheng2024evaluating,
  title={Evaluating Uncertainty-based Failure Detection for Closed-Loop LLM Planners},
  author={Zheng, Zhi and Feng, Qian and Li, Hang and Knoll, Alois and Feng, Jianxiang},
  booktitle=icra,
  year={2024}
}

@inproceedings{dai2025racer,
  title={Racer: Rich language-guided failure recovery policies for imitation learning},
  author={Dai, Yinpei and Lee, Jayjun and Fazeli, Nima and Chai, Joyce},
  booktitle=icra,
  year={2025}
}

@inproceedings{sogi2024future,
  title={Future predictive success-or-failure classification for long-horizon robotic tasks},
  author={Sogi, Naoya and Oyama, Hiroyuki and Shibata, Takashi and Terao, Makoto},
  booktitle={International Joint Conference on Neural Networks (IJCNN)},
  year={2024}
}

@inproceedings{
    miller2025counterfactual,
    title={Counterfactual reasoning: an analysis of in-context emergence},
    author={Moritz Miller and Bernhard Sch{\"o}lkopf and Siyuan Guo},
    booktitle=neurips,
    year={2025}
}

@inproceedings{mandlekar2022matters,
  title={What Matters in Learning from Offline Human Demonstrations for Robot Manipulation},
  author={Mandlekar, Ajay and Xu, Danfei and Wong, Josiah and Nasiriany, Soroush and Wang, Chen and Kulkarni, Rohun and Fei-Fei, Li and Savarese, Silvio and Zhu, Yuke and Mart{\'\i}n-Mart{\'\i}n, Roberto},
  booktitle=corl,
  year={2022}
}

@inproceedings{
    jia2024towards,
    title={Towards Diverse Behaviors: A Benchmark for Imitation Learning with Human Demonstrations},
    author={Xiaogang Jia and Denis Blessing and Xinkai Jiang and Moritz Reuss and Atalay Donat and Rudolf Lioutikov and Gerhard Neumann},
    booktitle=iclr,
    year={2024}
}

@article{chi2025diffusion,
  title={Diffusion policy: Visuomotor policy learning via action diffusion},
  author={Chi, Cheng and Xu, Zhenjia and Feng, Siyuan and Cousineau, Eric and Du, Yilun and Burchfiel, Benjamin and Tedrake, Russ and Song, Shuran},
  journal=ijrr,
  year={2025}
}

@inproceedings{agia2025unpacking,
  title={Unpacking Failure Modes of Generative Policies: Runtime Monitoring of Consistency and Progress},
  author={Agia, Christopher and Sinha, Rohan and Yang, Jingyun and Cao, Ziang and Antonova, Rika and Pavone, Marco and Bohg, Jeannette},
  booktitle=corl,
  year={2025}
}

@inproceedings{lipman2023flow,
  title={Flow Matching for Generative Modeling},
  author={Lipman, Yaron and Chen, Ricky TQ and Ben-Hamu, Heli and Nickel, Maximilian and Le, Matt},
  booktitle=iclr,
  year={2023}
}

@article{lamb2016professor,
  title={Professor forcing: A new algorithm for training recurrent networks},
  author={Lamb, Alex M and Alias Parth Goyal, Anirudh Goyal and Zhang, Ying and Zhang, Saizheng and Courville, Aaron C and Bengio, Yoshua},
  journal=neurips,
  year={2016}
}

@inproceedings{besl1992method,
  title={Method for registration of 3-D shapes},
  author={Besl, Paul J and McKay, Neil D},
  booktitle={Sensor fusion IV: control paradigms and data structures},
  year={1992}
}

@InProceedings{Yang_2013_ICCV,
    author = {Yang, Jiaolong and Li, Hongdong and Jia, Yunde},
    title = {Go-ICP: Solving 3D Registration Efficiently and Globally Optimally},
    booktitle = iccv,
    year = {2013}
}

@inproceedings{he2016deep,
  title={Deep residual learning for image recognition},
  author={He, Kaiming and Zhang, Xiangyu and Ren, Shaoqing and Sun, Jian},
  booktitle=cvpr,
  year={2016}
}

@article{mason1986mechanics,
  title={Mechanics and planning of manipulator pushing operations},
  author={Mason, Matthew T},
  journal={The International Journal of Robotics Research},
  year={1986}
}

@article{lynch1996stable,
  title={Stable pushing: Mechanics, controllability, and planning},
  author={Lynch, Kevin M and Mason, Matthew T},
  journal={The international journal of robotics research},
  year={1996}
}

@ARTICLE{8280543,
  author={Ruggiero, Fabio and Lippiello, Vincenzo and Siciliano, Bruno},
  journal=ieeera, 
  title={Nonprehensile Dynamic Manipulation: A Survey}, 
  year={2018}
}

\clearpage
\begin{strip}
\begin{center}
    {\Large \bfseries Appendix}
\end{center}
\vspace{1em}
\end{strip}

\section{Data Post-processing}

Our data post-processing involves a \textbf{hard cut-off}.
We remove time steps after $t_\text{fail}$. This is because the relative motion is minimal or absent both when the robot is moving steadily and after a failure, when the object remains stationary on the ground and the robot stays fixed. In such cases, the model may struggle to distinguish post-failure states from the initial low-risk stage due to similarly negligible relative motion. Furthermore, post-failure data is not informative for proactive failure prediction and is therefore excluded from training.

\section{Reliability of 2D ICP}

\begin{figure}[h]
    \centering
    \includegraphics[width=\linewidth]{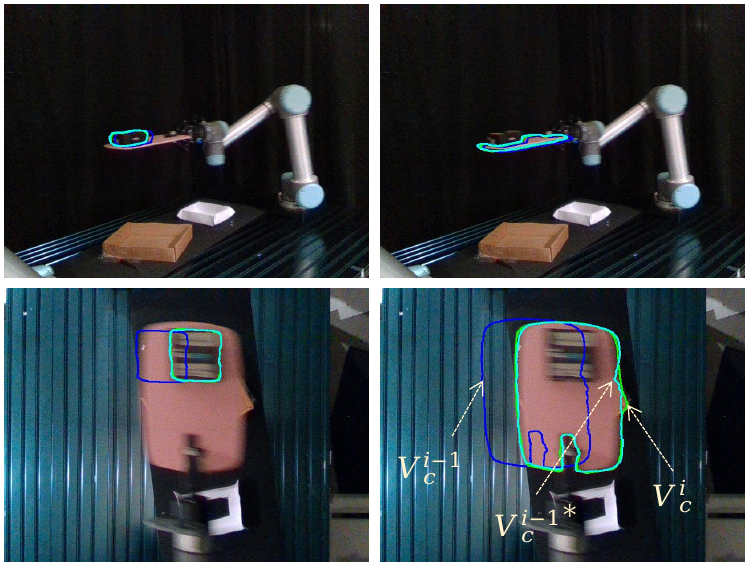}
    \caption{Reliability of 2D ICP. The contours from the last step are shown in blue, the contours from the current step are shown in green, and the contours obtained by applying rotation and translation relative to the last step are shown in cyan.}
    \label{fig:icp}
\end{figure}

We use 2D ICP to calculate the relative motions $M$ between the object and the carrier based on their contours $\{V_o^{i}, V_c^{i}\}_\text{top, side}$ in both camera views at each time step $i$. To verify the correctness and reliability of the calculated rotation and translation values, we visualize the contours and compare the calculated contour at $i+1$ time step with the ground truth, see Fig.~\ref{fig:icp}. The blue contours indicates $\{V_o^{i-1}, V_c^{i-1}\}_\text{top, side}$, while the green contours indicates $\{V_o^{i}, V_c^{i}\}_\text{top, side}$, and the cyan contours are $\{V_o^{i}, V_c^{i}\}_\text{top, side}^*$ transformed by the calculated 2D ICP (rotation and translation).

While we calculate the relative motion between the current step and the initial step, the motion values are accumulated across all intermediate adjacent steps. The accuracy of 2D ICP depends heavily on the precision of the extracted contours. In simulation, contours are obtained directly from object projections, whereas in the real world they are derived from SAM2-based segmentation. As a result, the relative motion computed from simulated data is naturally more precise than that from real-world data.

\section{Past Risk Data Augmentation}

In our network design, given a sampling window of size $m$, the robot state input branch includes the past risk values from the previous $m$ steps. During training, the model uses these ground-truth values for forward prediction, whereas during inference it must rely on the predicted risk values from the previous step. To reduce the discrepancy between training and inference, where ground-truth past risks are unavailable, we inject random noise into the ground-truth risk values during training so the model becomes robust to imperfect previous predictions, similar in spirit to techniques used to mitigate exposure bias in teacher forcing.

\section{Dataset}

Our collected dataset is stored in `.npy' files. For training, we split the data into training, validation, and test sets using an 8:1:1 ratio. During this process, we control that failure and non-failure roll-outs are evenly distributed across the three splits.

\begin{table}[t]
    \centering
	\resizebox{0.5\textwidth}{!}{
    \begin{tabular}{ccccc}
    \toprule
    Modules & Precision $\uparrow$ & Accuracy $\uparrow$ & Recall $\uparrow$ & $\eta \uparrow$ \\
    \midrule
    \textbf{Hermite (used)} & \textbf{0.995} & \textbf{0.997} & 0.995 & \textbf{0.923}\\
    \textbf{Bézier} & 0.973 & 0.988 & 0.995 & 0.887\\
    \textbf{Linear} & 0.958 & 0.980 & \textbf{1.0000} & 0.825\\
    \textbf{Step} & 0.314 & 0.314 & \textbf{1.0000} & 0.1373\\
    \bottomrule
    \end{tabular}
    }
    \caption{Comparison of different interpolation techniques.}
    \label{tab:splines}
    \vspace{-10pt}
\end{table}

\section{Simulation}

In simulation, we use a single UR5 robot with a flat panel attached to the end-effector as the carrier. For each roll-out, we randomize multiple aspects of the scene. For the panel, we randomize its color, surface friction, target tilt angle, rotation speed, and whether the panel reverses direction or holds upon reaching the target angle. For the object, we randomize its color (set as the inverse of the panel color to ensure visual contrast), size, mass, inertia, surface friction, and its initial position on the carrier; the object's initial orientation is sampled from a predefined configuration. For the robot, we randomize the execution trajectory by both shuffling the order of waypoints and applying per-joint angle perturbations, as well as independently randomizing the movement speed and the pause duration between waypoints. For the background, we randomize the ground texture (selected from a pool of materials), texture repeat scale, surface reflectance, and apply small RGB noise to the floor color. Finally, we introduce independent perturbations to the 3D positions of both the top-view and front-view cameras.

\section{Splines}

For the ground truth risk value in each failure roll-out. After defining the $t_0$, $t_\text{stop}$, and $t_\text{fail}$, we use a Hermite spline to interpolate the risk values for the remaining time steps. We also evaluate alternative interpolation methods, including Bézier interpolation, linear interpolation and step interpolation. The comparison results are summarized in Tab.~\ref{tab:splines}. With continuous interpolated values such as splines, the overall performance is significantly better than with step interpolation. This is consistent with our discussion in Sec. V-G.

\section{Complete Performance Report for Real-World Cross Validation}

Since we perform 10-fold cross-validation for the real-world experiments, we report the results from all folds (Tables~\ref{tab:experiment_real_split1}--\ref{tab:experiment_real_split10}) alongside the average performance across folds (Table~\ref{tab:experiment_real_avg}) to provide a comprehensive evaluation of the model's performance.

\clearpage

\begin{table*}[h]
    \centering
    \begin{tabular}{c|c|c|cccc}
    \toprule
    \multicolumn{2}{c|}{Methods / Variants} & Encoder Arch & Precision $\uparrow$ & Accuracy $\uparrow$ & Recall $\uparrow$ & $\eta \uparrow$ \\
    \midrule
    \multirow{10}{*}{Fail-Detect~\cite{XuC2-RSS-25}} & (CFM~\cite{yang2024consistency}, FM) & \multirow{10}{*}{ResNet18} & 0.481 & 0.481 & \textbf{1.000} & 0.377\\
    & (CFM~\cite{yang2024consistency}, DP) & & 0.481 & 0.481 & \textbf{1.000} & 0.359\\
    & (NatPN~\cite{charpentier2022natural}, FM) & & 0.481 & 0.481 & \textbf{1.000} & 0.311\\
    & (NatPN~\cite{charpentier2022natural}, DP) & & 0.481 & 0.481 & \textbf{1.000} & 0.311\\
    & (PCA-kmeans~\cite{liu2025multi}, FM) & & 0.481 & 0.481 & \textbf{1.000} & 0.172\\
    & (PCA-kmeans~\cite{liu2025multi}, DP) & & 0.444 & 0.444 & \textbf{1.000} & 0.269\\
    & (RND~\cite{he2024rediffuser}, FM) & & 0.481 & 0.481 & \textbf{1.000} & 0.329\\
    & (RND~\cite{he2024rediffuser}, DP) & & 0.481 & 0.481 & \textbf{1.000} & 0.367\\
    & (logpZO~\cite{XuC2-RSS-25}, FM) & & 0.481 & 0.481 & \textbf{1.000} & 0.231\\
    & (logpZO~\cite{XuC2-RSS-25}, DP) & & 0.481 & 0.481 & \textbf{1.000} & 0.227\\
    \midrule
    \multicolumn{2}{c|}{\textbf{\method (ours)}} & ResNet18 & \textbf{0.867} & \textbf{0.926} & \textbf{1.000} & \textbf{0.943}\\
    \multicolumn{2}{c|}{\textbf{\method (ours)}} & ResNet50 & 0.800 & 0.889 & \textbf{1.000} & 0.927\\
    \multicolumn{2}{c|}{\textbf{\method (ours)}} & ResNet152 & 0.857 & 0.926 & \textbf{1.000} & 0.969\\
    \bottomrule
    \end{tabular}
    \caption{Performance on real-world data cross validation split 1.}
    \label{tab:experiment_real_split1}
\end{table*}
 
\begin{table*}[h]
    \centering
    \begin{tabular}{c|c|c|cccc}
    \toprule
    \multicolumn{2}{c|}{Methods / Variants} & Encoder Arch & Precision $\uparrow$ & Accuracy $\uparrow$ & Recall $\uparrow$ & $\eta \uparrow$ \\
    \midrule
    \multirow{10}{*}{Fail-Detect~\cite{XuC2-RSS-25}} & (CFM~\cite{yang2024consistency}, FM) & \multirow{10}{*}{ResNet18} & 0.519 & 0.519 & \textbf{1.000} & 0.270\\
    & (CFM~\cite{yang2024consistency}, DP) & & 0.519 & 0.519 & \textbf{1.000} & 0.205\\
    & (NatPN~\cite{charpentier2022natural}, FM) & & 0.519 & 0.519 & \textbf{1.000} & 0.211\\
    & (NatPN~\cite{charpentier2022natural}, DP) & & 0.519 & 0.519 & \textbf{1.000} & 0.211\\
    & (PCA-kmeans~\cite{liu2025multi}, FM) & & 0.519 & 0.519 & \textbf{1.000} & 0.106\\
    & (PCA-kmeans~\cite{liu2025multi}, DP) & & 0.519 & 0.519 & \textbf{1.000} & 0.110\\
    & (RND~\cite{he2024rediffuser}, FM) & & 0.519 & 0.519 & \textbf{1.000} & 0.125\\
    & (RND~\cite{he2024rediffuser}, DP) & & 0.519 & 0.519 & \textbf{1.000} & 0.143\\
    & (logpZO~\cite{XuC2-RSS-25}, FM) & & 0.519 & 0.519 & \textbf{1.000} & 0.100\\
    & (logpZO~\cite{XuC2-RSS-25}, DP) & & 0.519 & 0.519 & \textbf{1.000} & 0.100\\
    \midrule
    \multicolumn{2}{c|}{\textbf{\method (ours)}} & ResNet18 & \textbf{1.000} & \textbf{1.000} & \textbf{1.000} & \textbf{0.943}\\
    \multicolumn{2}{c|}{\textbf{\method (ours)}} & ResNet50 & \textbf{1.000} & \textbf{1.000} & \textbf{1.000} & 0.904\\
    \multicolumn{2}{c|}{\textbf{\method (ours)}} & ResNet152 & 0.933 & 0.963 & \textbf{1.000} & 0.912\\
    \bottomrule
    \end{tabular}
    \caption{Performance on real-world data cross validation split 2.}
    \label{tab:experiment_real_split2}
\end{table*}
 
\begin{table*}[h]
    \centering
    \begin{tabular}{c|c|c|cccc}
    \toprule
    \multicolumn{2}{c|}{Methods / Variants} & Encoder Arch & Precision $\uparrow$ & Accuracy $\uparrow$ & Recall $\uparrow$ & $\eta \uparrow$ \\
    \midrule
    \multirow{10}{*}{Fail-Detect~\cite{XuC2-RSS-25}} & (CFM~\cite{yang2024consistency}, FM) & \multirow{10}{*}{ResNet18} & 0.519 & 0.519 & \textbf{1.000} & 0.184\\
    & (CFM~\cite{yang2024consistency}, DP) & & 0.519 & 0.519 & \textbf{1.000} & 0.346\\
    & (NatPN~\cite{charpentier2022natural}, FM) & & 0.481 & 0.481 & \textbf{1.000} & 0.159\\
    & (NatPN~\cite{charpentier2022natural}, DP) & & 0.481 & 0.481 & \textbf{1.000} & 0.159\\
    & (PCA-kmeans~\cite{liu2025multi}, FM) & & 0.519 & 0.519 & \textbf{1.000} & 0.283\\
    & (PCA-kmeans~\cite{liu2025multi}, DP) & & 0.519 & 0.519 & \textbf{1.000} & 0.097\\
    & (RND~\cite{he2024rediffuser}, FM) & & 0.519 & 0.519 & \textbf{1.000} & 0.366\\
    & (RND~\cite{he2024rediffuser}, DP) & & 0.519 & 0.519 & \textbf{1.000} & 0.248\\
    & (logpZO~\cite{XuC2-RSS-25}, FM) & & 0.519 & 0.519 & \textbf{1.000} & 0.196\\
    & (logpZO~\cite{XuC2-RSS-25}, DP) & & 0.519 & 0.519 & \textbf{1.000} & 0.241\\
    \midrule
    \multicolumn{2}{c|}{\textbf{\method (ours)}} & ResNet18 & \textbf{0.929} & \textbf{0.963} & \textbf{1.000} & \textbf{0.891}\\
    \multicolumn{2}{c|}{\textbf{\method (ours)}} & ResNet50 & \textbf{0.929} & \textbf{0.963} & \textbf{1.000} & 0.834\\
    \multicolumn{2}{c|}{\textbf{\method (ours)}} & ResNet152 & 0.857 & 0.926 & \textbf{1.000} & 0.896\\
    \bottomrule
    \end{tabular}
    \caption{Performance on real-world data cross validation split 3.}
    \label{tab:experiment_real_split3}
\end{table*}
 
\begin{table*}[h]
    \centering
    \begin{tabular}{c|c|c|cccc}
    \toprule
    \multicolumn{2}{c|}{Methods / Variants} & Encoder Arch & Precision $\uparrow$ & Accuracy $\uparrow$ & Recall $\uparrow$ & $\eta \uparrow$ \\
    \midrule
    \multirow{10}{*}{Fail-Detect~\cite{XuC2-RSS-25}} & (CFM~\cite{yang2024consistency}, FM) & \multirow{10}{*}{ResNet18} & 0.481 & 0.481 & \textbf{1.000} & 0.276\\
    & (CFM~\cite{yang2024consistency}, DP) & & 0.481 & 0.481 & \textbf{1.000} & 0.324\\
    & (NatPN~\cite{charpentier2022natural}, FM) & & 0.519 & 0.519 & \textbf{1.000} & 0.187\\
    & (NatPN~\cite{charpentier2022natural}, DP) & & 0.519 & 0.519 & \textbf{1.000} & 0.187\\
    & (PCA-kmeans~\cite{liu2025multi}, FM) & & 0.519 & 0.519 & \textbf{1.000} & 0.109\\
    & (PCA-kmeans~\cite{liu2025multi}, DP) & & 0.519 & 0.519 & \textbf{1.000} & 0.119\\
    & (RND~\cite{he2024rediffuser}, FM) & & 0.481 & 0.481 & \textbf{1.000} & 0.225\\
    & (RND~\cite{he2024rediffuser}, DP) & & 0.519 & 0.519 & \textbf{1.000} & 0.251\\
    & (logpZO~\cite{XuC2-RSS-25}, FM) & & 0.519 & 0.519 & \textbf{1.000} & 0.131\\
    & (logpZO~\cite{XuC2-RSS-25}, DP) & & 0.519 & 0.519 & \textbf{1.000} & 0.096\\
    \midrule
    \multicolumn{2}{c|}{\textbf{\method (ours)}} & ResNet18 & \textbf{0.929} & 0.926 & 0.929 & \textbf{0.929}\\
    \multicolumn{2}{c|}{\textbf{\method (ours)}} & ResNet50 & 0.867 & \textbf{0.926} & \textbf{1.000} & 0.908\\
    \multicolumn{2}{c|}{\textbf{\method (ours)}} & ResNet152 & \textbf{0.929} & \textbf{0.926} & 0.929 & 0.910\\
    \bottomrule
    \end{tabular}
    \caption{Performance on real-world data cross validation split 4.}
    \label{tab:experiment_real_split4}
\end{table*}
 
\begin{table*}[h]
    \centering
    \begin{tabular}{c|c|c|cccc}
    \toprule
    \multicolumn{2}{c|}{Methods / Variants} & Encoder Arch & Precision $\uparrow$ & Accuracy $\uparrow$ & Recall $\uparrow$ & $\eta \uparrow$ \\
    \midrule
    \multirow{10}{*}{Fail-Detect~\cite{XuC2-RSS-25}} & (CFM~\cite{yang2024consistency}, FM) & \multirow{10}{*}{ResNet18} & 0.407 & 0.407 & \textbf{1.000} & 0.367\\
    & (CFM~\cite{yang2024consistency}, DP) & & 0.481 & 0.481 & \textbf{1.000} & 0.411\\
    & (NatPN~\cite{charpentier2022natural}, FM) & & 0.407 & 0.407 & \textbf{1.000} & 0.290\\
    & (NatPN~\cite{charpentier2022natural}, DP) & & 0.407 & 0.407 & \textbf{1.000} & 0.290\\
    & (PCA-kmeans~\cite{liu2025multi}, FM) & & 0.444 & 0.444 & \textbf{1.000} & 0.249\\
    & (PCA-kmeans~\cite{liu2025multi}, DP) & & 0.444 & 0.444 & \textbf{1.000} & 0.312\\
    & (RND~\cite{he2024rediffuser}, FM) & & 0.519 & 0.519 & \textbf{1.000} & 0.394\\
    & (RND~\cite{he2024rediffuser}, DP) & & 0.519 & 0.519 & \textbf{1.000} & 0.348\\
    & (logpZO~\cite{XuC2-RSS-25}, FM) & & 0.519 & 0.519 & \textbf{1.000} & 0.185\\
    & (logpZO~\cite{XuC2-RSS-25}, DP) & & 0.481 & 0.481 & \textbf{1.000} & 0.332\\
    \midrule
    \multicolumn{2}{c|}{\textbf{\method (ours)}} & ResNet18 & \textbf{1.000} & \textbf{1.000} & \textbf{1.000} & \textbf{0.936}\\
    \multicolumn{2}{c|}{\textbf{\method (ours)}} & ResNet50 & 0.929 & 0.963 & \textbf{1.000} & 0.932\\
    \multicolumn{2}{c|}{\textbf{\method (ours)}} & ResNet152 & 0.929 & 0.963 & \textbf{1.000} & 0.967\\
    \bottomrule
    \end{tabular}
    \caption{Performance on real-world data cross validation split 5.}
    \label{tab:experiment_real_split5}
\end{table*}
 
\begin{table*}[h]
    \centering
    \begin{tabular}{c|c|c|cccc}
    \toprule
    \multicolumn{2}{c|}{Methods / Variants} & Encoder Arch & Precision $\uparrow$ & Accuracy $\uparrow$ & Recall $\uparrow$ & $\eta \uparrow$ \\
    \midrule
    \multirow{10}{*}{Fail-Detect~\cite{XuC2-RSS-25}} & (CFM~\cite{yang2024consistency}, FM) & \multirow{10}{*}{ResNet18} & 0.481 & 0.481 & \textbf{1.000} & 0.337\\
    & (CFM~\cite{yang2024consistency}, DP) & & 0.519 & 0.519 & \textbf{1.000} & 0.296\\
    & (NatPN~\cite{charpentier2022natural}, FM) & & 0.519 & 0.519 & \textbf{1.000} & 0.403\\
    & (NatPN~\cite{charpentier2022natural}, DP) & & 0.519 & 0.519 & \textbf{1.000} & 0.403\\
    & (PCA-kmeans~\cite{liu2025multi}, FM) & & 0.481 & 0.481 & \textbf{1.000} & 0.345\\
    & (PCA-kmeans~\cite{liu2025multi}, DP) & & 0.481 & 0.481 & \textbf{1.000} & 0.413\\
    & (RND~\cite{he2024rediffuser}, FM) & & 0.519 & 0.519 & \textbf{1.000} & 0.300\\
    & (RND~\cite{he2024rediffuser}, DP) & & 0.519 & 0.519 & \textbf{1.000} & 0.367\\
    & (logpZO~\cite{XuC2-RSS-25}, FM) & & 0.519 & 0.519 & \textbf{1.000} & 0.273\\
    & (logpZO~\cite{XuC2-RSS-25}, DP) & & 0.519 & 0.519 & \textbf{1.000} & 0.274\\
    \midrule
    \multicolumn{2}{c|}{\textbf{\method (ours)}} & ResNet18 & \textbf{0.929} & \textbf{0.963} & \textbf{1.000} & \textbf{0.932}\\
    \multicolumn{2}{c|}{\textbf{\method (ours)}} & ResNet50 & 0.846 & 0.889 & 0.917 & \textbf{0.938}\\
    \multicolumn{2}{c|}{\textbf{\method (ours)}} & ResNet152 & 0.923 & 0.926 & 0.923 & 0.936\\
    \bottomrule
    \end{tabular}
    \caption{Performance on real-world data cross validation split 6.}
    \label{tab:experiment_real_split6}
\end{table*}
 
\begin{table*}[h]
    \centering
    \begin{tabular}{c|c|c|cccc}
    \toprule
    \multicolumn{2}{c|}{Methods / Variants} & Encoder Arch & Precision $\uparrow$ & Accuracy $\uparrow$ & Recall $\uparrow$ & $\eta \uparrow$ \\
    \midrule
    \multirow{10}{*}{Fail-Detect~\cite{XuC2-RSS-25}} & (CFM~\cite{yang2024consistency}, FM) & \multirow{10}{*}{ResNet18} & 0.444 & 0.444 & \textbf{1.000} & 0.274\\
    & (CFM~\cite{yang2024consistency}, DP) & & 0.519 & 0.519 & \textbf{1.000} & 0.264\\
    & (NatPN~\cite{charpentier2022natural}, FM) & & 0.519 & 0.519 & \textbf{1.000} & 0.262\\
    & (NatPN~\cite{charpentier2022natural}, DP) & & 0.519 & 0.519 & \textbf{1.000} & 0.262\\
    & (PCA-kmeans~\cite{liu2025multi}, FM) & & 0.444 & 0.444 & \textbf{1.000} & 0.132\\
    & (PCA-kmeans~\cite{liu2025multi}, DP) & & 0.481 & 0.481 & \textbf{1.000} & 0.221\\
    & (RND~\cite{he2024rediffuser}, FM) & & 0.481 & 0.481 & \textbf{1.000} & 0.303\\
    & (RND~\cite{he2024rediffuser}, DP) & & 0.481 & 0.481 & \textbf{1.000} & 0.317\\
    & (logpZO~\cite{XuC2-RSS-25}, FM) & & 0.444 & 0.444 & \textbf{1.000} & 0.252\\
    & (logpZO~\cite{XuC2-RSS-25}, DP) & & 0.481 & 0.481 & \textbf{1.000} & 0.190\\
    \midrule
    \multicolumn{2}{c|}{\textbf{\method (ours)}} & ResNet18 & \textbf{1.000} & \textbf{1.000} & \textbf{1.000} & \textbf{0.940}\\
    \multicolumn{2}{c|}{\textbf{\method (ours)}} & ResNet50 & \textbf{1.000} & \textbf{1.000} & \textbf{1.000} & 0.828\\
    \multicolumn{2}{c|}{\textbf{\method (ours)}} & ResNet152 & 0.933 & 0.963 & \textbf{1.000} & 0.880\\
    \bottomrule
    \end{tabular}
    \caption{Performance on real-world data cross validation split 7.}
    \label{tab:experiment_real_split7}
\end{table*}
 
\begin{table*}[h]
    \centering
    \begin{tabular}{c|c|c|cccc}
    \toprule
    \multicolumn{2}{c|}{Methods / Variants} & Encoder Arch & Precision $\uparrow$ & Accuracy $\uparrow$ & Recall $\uparrow$ & $\eta \uparrow$ \\
    \midrule
    \multirow{10}{*}{Fail-Detect~\cite{XuC2-RSS-25}} & (CFM~\cite{yang2024consistency}, FM) & \multirow{10}{*}{ResNet18} & 0.519 & 0.519 & \textbf{1.000} & 0.291\\
    & (CFM~\cite{yang2024consistency}, DP) & & 0.519 & 0.519 & \textbf{1.000} & 0.274\\
    & (NatPN~\cite{charpentier2022natural}, FM) & & 0.519 & 0.519 & \textbf{1.000} & 0.208\\
    & (NatPN~\cite{charpentier2022natural}, DP) & & 0.519 & 0.519 & \textbf{1.000} & 0.208\\
    & (PCA-kmeans~\cite{liu2025multi}, FM) & & 0.519 & 0.519 & \textbf{1.000} & 0.201\\
    & (PCA-kmeans~\cite{liu2025multi}, DP) & & 0.519 & 0.519 & \textbf{1.000} & 0.184\\
    & (RND~\cite{he2024rediffuser}, FM) & & 0.519 & 0.519 & \textbf{1.000} & 0.208\\
    & (RND~\cite{he2024rediffuser}, DP) & & 0.519 & 0.519 & \textbf{1.000} & 0.288\\
    & (logpZO~\cite{XuC2-RSS-25}, FM) & & 0.519 & 0.519 & \textbf{1.000} & 0.148\\
    & (logpZO~\cite{XuC2-RSS-25}, DP) & & 0.519 & 0.519 & \textbf{1.000} & 0.182\\
    \midrule
    \multicolumn{2}{c|}{\textbf{\method (ours)}} & ResNet18 & \textbf{0.933} & \textbf{0.963} & \textbf{1.000} & \textbf{0.958}\\
    \multicolumn{2}{c|}{\textbf{\method (ours)}} & ResNet50 & \textbf{1.000} & \textbf{1.000} & \textbf{1.000} & 0.946\\
    \multicolumn{2}{c|}{\textbf{\method (ours)}} & ResNet152 & 0.778 & 0.852 & \textbf{1.000} & 0.949\\
    \bottomrule
    \end{tabular}
    \caption{Performance on real-world data cross validation split 8.}
    \label{tab:experiment_real_split8}
\end{table*}
 
\begin{table*}[h]
    \centering
    \begin{tabular}{c|c|c|cccc}
    \toprule
    \multicolumn{2}{c|}{Methods / Variants} & Encoder Arch & Precision $\uparrow$ & Accuracy $\uparrow$ & Recall $\uparrow$ & $\eta \uparrow$ \\
    \midrule
    \multirow{10}{*}{Fail-Detect~\cite{XuC2-RSS-25}} & (CFM~\cite{yang2024consistency}, FM) & \multirow{10}{*}{ResNet18} & 0.444 & 0.444 & \textbf{1.000} & 0.329\\
    & (CFM~\cite{yang2024consistency}, DP) & & 0.481 & 0.481 & \textbf{1.000} & 0.201\\
    & (NatPN~\cite{charpentier2022natural}, FM) & & 0.519 & 0.519 & \textbf{1.000} & 0.366\\
    & (NatPN~\cite{charpentier2022natural}, DP) & & 0.519 & 0.519 & \textbf{1.000} & 0.366\\
    & (PCA-kmeans~\cite{liu2025multi}, FM) & & 0.481 & 0.481 & \textbf{1.000} & 0.169\\
    & (PCA-kmeans~\cite{liu2025multi}, DP) & & 0.444 & 0.444 & \textbf{1.000} & 0.174\\
    & (RND~\cite{he2024rediffuser}, FM) & & 0.481 & 0.481 & \textbf{1.000} & 0.420\\
    & (RND~\cite{he2024rediffuser}, DP) & & 0.519 & 0.519 & \textbf{1.000} & 0.377\\
    & (logpZO~\cite{XuC2-RSS-25}, FM) & & 0.519 & 0.519 & \textbf{1.000} & 0.061\\
    & (logpZO~\cite{XuC2-RSS-25}, DP) & & 0.444 & 0.444 & \textbf{1.000} & 0.151\\
    \midrule
    \multicolumn{2}{c|}{\textbf{\method (ours)}} & ResNet18 & \textbf{0.933} & \textbf{0.963} & \textbf{1.000} & \textbf{0.915}\\
    \multicolumn{2}{c|}{\textbf{\method (ours)}} & ResNet50 & \textbf{0.929} & \textbf{0.963} & \textbf{1.000} & \textbf{0.931}\\
    \multicolumn{2}{c|}{\textbf{\method (ours)}} & ResNet152 & \textbf{0.933} & \textbf{0.963} & \textbf{1.000} & 0.908\\
    \bottomrule
    \end{tabular}
    \caption{Performance on real-world data cross validation split 9.}
    \label{tab:experiment_real_split9}
\end{table*}
 
\begin{table*}[h]
    \centering
    \begin{tabular}{c|c|c|cccc}
    \toprule
    \multicolumn{2}{c|}{Methods / Variants} & Encoder Arch & Precision $\uparrow$ & Accuracy $\uparrow$ & Recall $\uparrow$ & $\eta \uparrow$ \\
    \midrule
    \multirow{10}{*}{Fail-Detect~\cite{XuC2-RSS-25}} & (CFM~\cite{yang2024consistency}, FM) & \multirow{10}{*}{ResNet18} & 0.481 & 0.481 & \textbf{1.000} & 0.267\\
    & (CFM~\cite{yang2024consistency}, DP) & & 0.481 & 0.481 & \textbf{1.000} & 0.345\\
    & (NatPN~\cite{charpentier2022natural}, FM) & & 0.444 & 0.444 & \textbf{1.000} & 0.355\\
    & (NatPN~\cite{charpentier2022natural}, DP) & & 0.444 & 0.444 & \textbf{1.000} & 0.355\\
    & (PCA-kmeans~\cite{liu2025multi}, FM) & & 0.519 & 0.519 & \textbf{1.000} & 0.495\\
    & (PCA-kmeans~\cite{liu2025multi}, DP) & & 0.519 & 0.519 & \textbf{1.000} & 0.305\\
    & (RND~\cite{he2024rediffuser}, FM) & & 0.519 & 0.519 & \textbf{1.000} & 0.218\\
    & (RND~\cite{he2024rediffuser}, DP) & & 0.481 & 0.481 & \textbf{1.000} & 0.314\\
    & (logpZO~\cite{XuC2-RSS-25}, FM) & & 0.481 & 0.481 & \textbf{1.000} & 0.370\\
    & (logpZO~\cite{XuC2-RSS-25}, DP) & & 0.481 & 0.481 & \textbf{1.000} & 0.396\\
    \midrule
    \multicolumn{2}{c|}{\textbf{\method (ours)}} & ResNet18 & \textbf{0.923} & \textbf{0.926} & 0.923 & \textbf{0.937}\\
    \multicolumn{2}{c|}{\textbf{\method (ours)}} & ResNet50 & 0.857 & 0.889 & 0.923 & \textbf{0.957}\\
    \multicolumn{2}{c|}{\textbf{\method (ours)}} & ResNet152 & \textbf{0.923} & \textbf{0.926} & 0.923 & 0.922\\
    \bottomrule
    \end{tabular}
    \caption{Performance on real-world data cross validation split 10.}
    \label{tab:experiment_real_split10}
\end{table*}
 
\begin{table*}[h]
    \centering
    \begin{tabular}{c|c|c|cccc}
    \toprule
    \multicolumn{2}{c|}{Methods / Variants} & Encoder Arch & Precision $\uparrow$ & Accuracy $\uparrow$ & Recall $\uparrow$ & $\eta \uparrow$ \\
    \midrule
    \multirow{10}{*}{Fail-Detect~\cite{XuC2-RSS-25}} & (CFM~\cite{yang2024consistency}, FM) & \multirow{10}{*}{ResNet18} & 0.478 & 0.478 & \textbf{1.000} & 0.297\\
    & (CFM~\cite{yang2024consistency}, DP) & & 0.500 & 0.500 & \textbf{1.000} & 0.303\\
    & (NatPN~\cite{charpentier2022natural}, FM) & & 0.493 & 0.493 & \textbf{1.000} & 0.275\\
    & (NatPN~\cite{charpentier2022natural}, DP) & & 0.493 & 0.493 & \textbf{1.000} & 0.275\\
    & (PCA-kmeans~\cite{liu2025multi}, FM) & & 0.493 & 0.493 & \textbf{1.000} & 0.226\\
    & (PCA-kmeans~\cite{liu2025multi}, DP) & & 0.489 & 0.489 & \textbf{1.000} & 0.220\\
    & (RND~\cite{he2024rediffuser}, FM) & & 0.504 & 0.504 & \textbf{1.000} & 0.289\\
    & (RND~\cite{he2024rediffuser}, DP) & & 0.507 & 0.507 & \textbf{1.000} & 0.302\\
    & (logpZO~\cite{XuC2-RSS-25}, FM) & & 0.504 & 0.504 & \textbf{1.000} & 0.195\\
    & (logpZO~\cite{XuC2-RSS-25}, DP) & & 0.496 & 0.496 & \textbf{1.000} & 0.219\\
    \midrule
    \multicolumn{2}{c|}{\textbf{\method (ours)}} & ResNet18 & \textbf{0.944} & \textbf{0.963} & 0.985 & \textbf{0.932}\\
    \multicolumn{2}{c|}{\textbf{\method (ours)}} & ResNet50 & 0.916 & 0.948 & 0.984 & 0.910\\
    \multicolumn{2}{c|}{\textbf{\method (ours)}} & ResNet152 & 0.900 & 0.933 & 0.977 & 0.925\\
    \bottomrule
    \end{tabular}
    \caption{Average performance on real-world data across all cross validation splits.}
    \label{tab:experiment_real_avg}
\end{table*}

\end{document}